\documentclass[runningheads]{llncs}

\usepackage{eccv}

\newcommand{\ours}{\texttt{ImAgent}\xspace}

\usepackage{eccvabbrv}

\usepackage{graphicx}
\usepackage{booktabs}
\usepackage[framemethod=TikZ]{mdframed}
\usepackage{tcolorbox}
\usepackage{enumitem}
\usepackage{multirow}
\usepackage{bigstrut}
\usepackage{xspace}

\usepackage[ruled,vlined]{algorithm2e}
\usepackage[accsupp]{axessibility}  

\usepackage{tcolorbox}
\tcbuselibrary{skins}

\usepackage[pagebackref,breaklinks,colorlinks,citecolor=eccvblue]{hyperref}

\usepackage{orcidlink}

\begin{document}


\title{\ours: A Unified Multimodal Agent Framework for Test-Time Scalable Image Generation}

\titlerunning{\ours for Image Generation}


\author{
Kaishen Wang$^{*}$, \quad
Ruibo Chen$^{*}$, \quad
Tong Zheng, \quad
Heng Huang
}

\authorrunning{K. Wang et al.}

\institute{University of Maryland, College Park, MD 20742, USA \\
\email{kaishen@umd.edu}}

\maketitle

\renewcommand{\thefootnote}{\fnsymbol{footnote}}
\footnotetext[1]{Equal Contribution}

\begin{figure*}[t]
    \centering
    \includegraphics[width=0.98\linewidth]{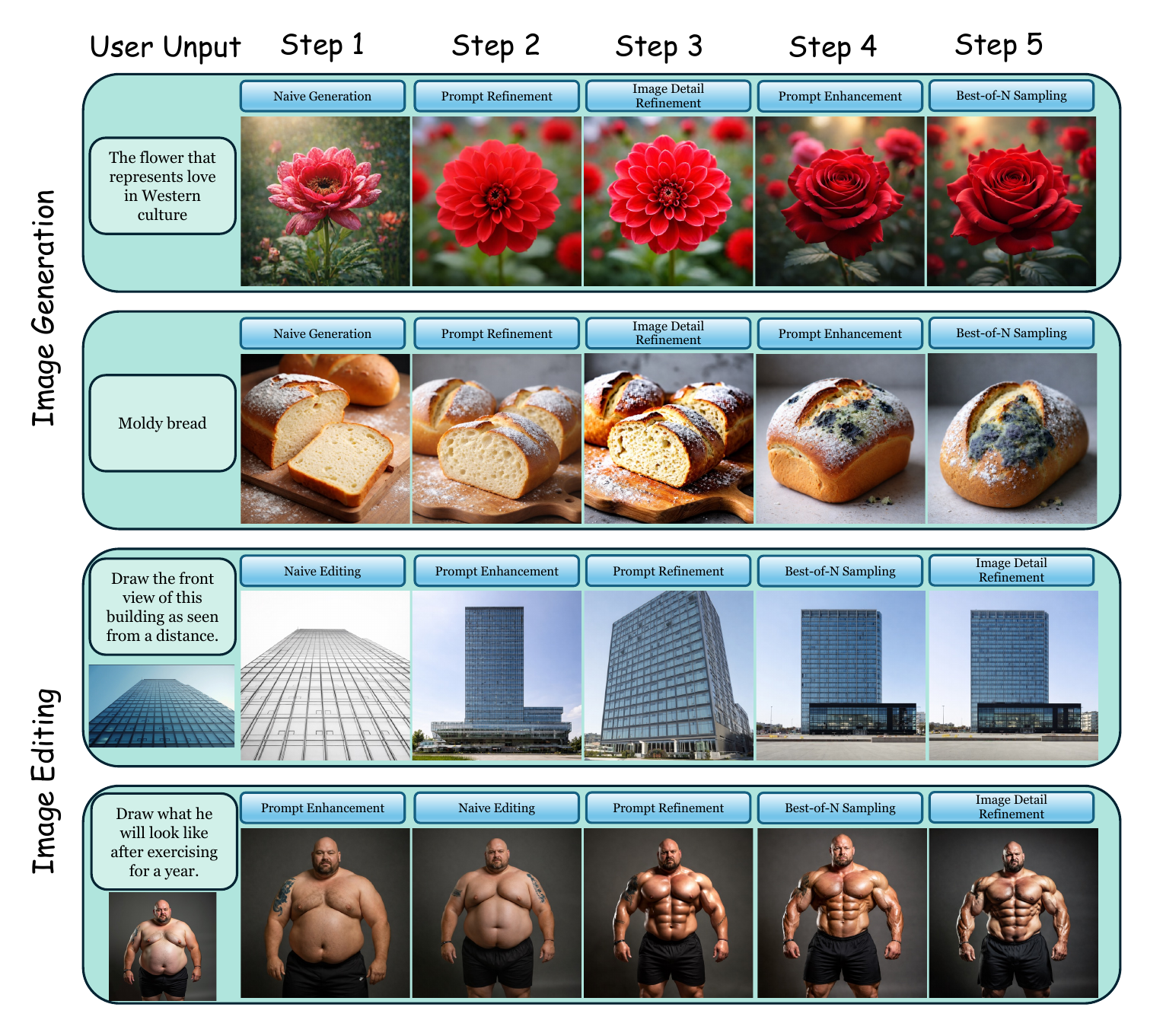}
    \vspace{-4mm}
    \caption{Qualitative examples of \ours built upon Bagel for image generation and image editing.}
    \label{fig:example}
    \vspace{-6mm}
\end{figure*}

\begin{abstract}
Recent text-to-image (T2I) models have achieved remarkable progress in generating visually realistic and semantically coherent images. However, their outputs often suffer from randomness and inconsistency with the given prompts, especially when textual descriptions are vague or underspecified. Existing test-time strategies, such as prompt rewriting, best-of-N sampling, and iterative refinement, can alleviate these issues but typically rely on multiple external components and operate independently, resulting in extra computational overhead and limited efficiency in test-time scaling. To address these challenges, we propose \ours, a training-free unified multimodal agent built upon Unified Multimodal Models (UMMs) for adaptive and efficient test-time scaling. Unlike prior pipelines that assemble separate models for reasoning, generation, and evaluation, \ours\ integrates a policy controller and multiple generation actions within a single UMM. The policy controller dynamically selects and coordinates actions based on observation history, enabling structured refinement and improved semantic alignment without introducing external modules. Extensive experiments on image generation and image editing tasks demonstrate that \ours\ consistently improves over its backbone model and even surpasses strong baselines in challenging cases where the backbone alone fails, highlighting the effectiveness of \ours for scalable and adaptive image generation.

\end{abstract}

\section{Introduction}
\label{sec:intro}
Text-to-Image (T2I) models~\cite{rombach2022high,ramesh2022hierarchical,podell2023sdxl,sun2024autoregressive,tian2024visual,qin2025lumina} have made remarkable progress in generating visually realistic and semantically coherent images from natural language descriptions. Despite these advancements, the quality of generated results often exhibits randomness and inconsistency with the given prompts~\cite{xu2024prompt,lee2024direct}. This limitation primarily stems from the strong dependency of existing T2I models on the clarity and specificity of textual inputs. When the prompt is vague or underspecified, the models tend to overlook key semantic elements, leading to images that deviate from the intended meaning and fail to fully capture the user’s intent.

To mitigate this important problem, prior research has proposed various strategies, such as prompt rewriting~\cite{brade2023promptify,mo2024dynamic,liu2025one}, best-of-N sampling~\cite{li2025reflect,singhal2025general}, classifier-free guidance~\cite{ho2022classifier,sauer2023stylegan,shen2024rethinking}, and self-revision or iterative refinement~\cite{singh2023divide,li2024g,jeon2025iterative}. These approaches aim to either clarify the textual prompt or reduce the model’s inherent randomness rather than finetuning the models, reflecting the underlying principle of test-time scaling—improving generation quality through additional inference-time computation.

While effective, these methods typically require additional components, such as a language model for prompt enhancement, a generative model for image synthesis, and a vision-language model for evaluation, thereby increasing memory consumption and computational cost. Moreover, they are often applied independently, requiring human intervention to determine the optimal approach for a given case, which substantially limits the efficiency of test-time scaling and leads to unnecessary computational overhead.

In this paper, we aim to construct a universal agent for image generation that can adaptively select the optimal action for a given case, allocate computational resources accordingly, and execute the selected action within the agent itself without relying on any external models. This design enables more efficient test-time scaling. Thanks to recent advances in unified multimodal models (UMMs)~\cite{deng2025emerging,wu2025janus,xie2024show,qin2025lumina}, which integrate text generation, image generation, and visual understanding within a single framework, this assumption becomes attainable. Building upon this foundation, we introduce \ours, a training-free unified multimodal agent designed to perform efficient test-time scaling for image generation.

Specifically, \ours\ is built around a policy controller that serves as its ``brain'', determining whether an image requires refinement based on the observation history and deciding which action should be invoked in the next step. Under its coordination, multiple predefined generation actions, such as \textit{Prompt Enhancement with CoT}, \textit{Image Detail Refinement}, and \textit{Best-of-N Sampling}, operate collaboratively within a unified framework. Through this dynamic interaction, \ours\ transforms what was previously a manually assembled pipeline into a self-organizing agent capable of reasoning, generation, and self-improvement, thereby enhancing the efficiency of test-time scaling. Notably, both the policy controller and all action executions are implemented entirely within the same unified model, without introducing any external modules or auxiliary networks.

We conduct experiments on two popular tasks, image generation and image editing, to evaluate the effectiveness of the proposed \ours. Extensive experimental results show that \ours\ achieves significant improvements over the backbone model, demonstrating its strong capability and the effectiveness of test-time scaling in enhancing image generation performance. The contributions of this paper can be summarized as follows:
\vspace{-1mm}
\begin{itemize}
    \item[(1)] We propose a new unified multimodal agent framework for image generation, \ours, which integrates multiple generation actions and dynamically coordinates them through a policy controller, where both the controller and all action executions are implemented within the same unified model without introducing external modules.
    \item[(2)] Extensive experiments on image generation and image editing tasks demonstrate that \ours\ achieves substantial improvements over the backbone model, even outperforming baselines that the backbone model fails to surpass, validating its effectiveness and in test-time scaling.
\end{itemize}

\section{Related Work}

\vspace{-1.8mm}
\subsection{Text-to-Image Generation}
Text-to-Image (T2I) models have demonstrated remarkable ability in synthesizing visually realistic and semantically coherent images from natural language descriptions. The evolution of T2I models has followed several major paradigms, including diffusion-based models~\cite{podell2023sdxl,rombach2022high,ramesh2022hierarchical,esser2024scaling,gao2024lumina,xie2025sana}, autoregressive models~\cite{sun2024autoregressive,zhang2024var,tian2024visual,wang2024emu3,zheng2025parallel,xiong2025llava}, and more recently, unified multimodal models (UMMs)~\cite{xiao2025omnigen,xie2024show,zhou2024transfusion,tong2025metamorph,deng2025emerging,zheng2025learning,luo2024stable}. These approaches have significantly advanced image fidelity, diversity, and text-image alignment.

\vspace{-3mm}
\subsection{Unified Multimodal Models}
UMMs refer to models that jointly possess multimodal understanding and generation capabilities within a single framework. Unlike traditional pipelines that rely on separate models for text or image understanding, image generation, and image editing, these models integrate multimodal understanding and image generation into a shared architecture, enabling seamless information exchange across modalities~\cite{xiao2025omnigen,zhou2024transfusion,tong2025metamorph,wu2025janus,wu2025janus}. This convergence of understanding and generation not only improves efficiency and coherence but also paves the way for building more general-purpose, human-like agents capable of reasoning, creating, and self-correcting.

\vspace{-2.5mm}
\subsection{Generation Optimization Strategies}
Despite the remarkable progress of T2I models, the quality of generated images still depends heavily on the clarity of prompts and the randomness inherent in the generation process. To alleviate these issues, various optimization strategies have been proposed. \textit{Best-of-N sampling}~\cite{tian2025unigen,verdun2025soft,li2025reflect} generates multiple candidates and selects the best one, reducing the influence of stochastic variations in model sampling. \textit{Prompt rewriting}~\cite{tian2025unigen,wang2024promptcharm,yasunaga2022retrieval} reformulates or enriches the input text to provide more explicit guidance for the generator, improving semantic alignment and visual coherence. \textit{Classifier-free guidance}~\cite{ho2022classifier,sauer2023stylegan,shen2024rethinking} controls the conditioning strength to balance fidelity and diversity, and \textit{iterative self-refinement}~\cite{yang2023idea2img,madaan2023self,singh2023divide,li2024g,jeon2025iterative} enables models to assess and revise their outputs through multiple rounds of generation. Although these techniques significantly enhance performance, they typically operate as independent modules with limited interaction between reasoning, generation, and evaluation, highlighting the need for unified multimodal frameworks that integrate these capabilities into a cohesive pipeline.

\vspace{-3mm}
\section{Proposed Method}

We create \ours for unified multimodal models (UMMs), which inherently possess both understanding and generation capabilities. Motivated by the observation that multimodal understanding is generally more reliable and easier to achieve than image generation~\cite{zhang2025unified,zhang2025unified1}, \ours leverages the model’s strong understanding ability to adaptively select the optimal action for a given case, then utilize the generation module to execute the corresponding action, thereby enhancing the overall generation quality. This design enables \ours to fully utilize the potential of a single UMM without any additional training or external modules, thereby achieving efficient test-time scaling. In this section, we first introduce the overall agent framework of \ours, followed by a detailed description of its action space. After that, we provide the details about the policy controller, including the specific prompt and action parsing.

\begin{figure*}[t]
    \centering
    \includegraphics[width=\linewidth]{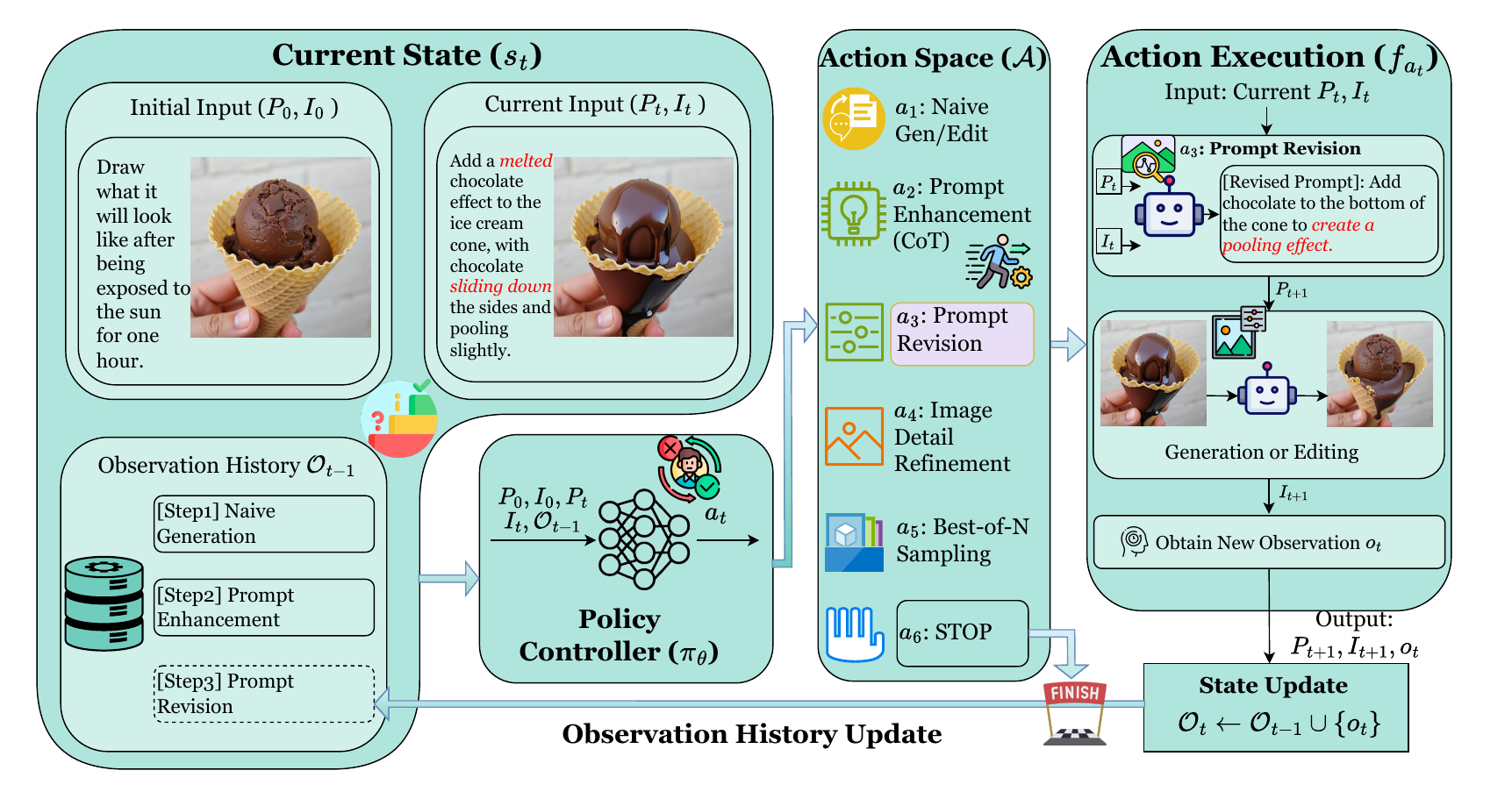}
    \vspace{-4mm}
    \caption{The overall architecture of our \ours framework.}
    \label{fig:overall-arch}
    \vspace{-4mm}
\end{figure*}

\vspace{-3mm}
\subsection{Agent Framework Overview}
As shown in Figure~\ref{fig:overall-arch}, \ours\ dynamically selects the next action based on the current observation and executes the selected action until either the \texttt{STOP} action is chosen or the maximum step limit is reached. Formally, we define the state at the current $t$-th time step as:
\vspace{-1.5mm}
\begin{equation}
s_t = \{P_0, I_0, P_t, I_t, \mathcal{O}_{t-1}\},
\vspace{-1.5mm}
\end{equation}
where $P_0$ and $I_0$ denote the initial user prompt and input image, $P_t$ and $I_t$ represent the current prompt and generated image at $t$-th step, and $\mathcal{O}_{t-1} = \{o_1, \dots, o_{t-1}\}$ denotes the history of past observations, i.e., action history and prompt history.

\ours dynamically selects an action $a_t$ from the defined action space $\mathcal{A} = \{a_1, \dots, a_n\}$ (detailed in Section~\ref{sec:action-space}) using a policy controller $\pi_\theta$ conditioned on the current state at $t$-th step:
\vspace{-1.5mm}
\begin{equation}
a_t \sim \pi_\theta(a \mid s_t).
\vspace{-1.5mm}
\end{equation}
Upon selecting $a_t$, the corresponding action function $f_{a_t}$ is invoked to produce an updated prompt $P_{t+1}$, an updated image $I_{t+1}$, and a new observation $o_t$:
\begin{equation}
(P_{t+1}, I_{t+1}, o_t) = f_{a_t}(P_t, I_t, \mathcal{O}_{t-1}).
\end{equation}
After that, the observation is updated as follows:
\begin{equation}
\mathcal{O}_t \leftarrow \mathcal{O}_{t-1} \cup {o_t}.
\end{equation}
Unlike conventional agent-based frameworks where observations directly influence state updates, in \ours\ the observation history is primarily used for action selection, while prompt and image updates are determined by the selected generation actions. 

This iterative process continues until the agent selects the \texttt{STOP} action, which indicates that the generated result is satisfactory, or until the maximum number of steps $T_{\max}$ is reached. Notably, both the policy controller and the action execution are implemented within a single UMM. The detailed algorithm for image generation is presented in Algorithm~\ref{alg:image_agent}, while the algorithm for image editing follows a similar procedure and is provided in the Appendix.

\begin{algorithm}[t]
\caption{\ours for Image Generation.}
\label{alg:image_agent}
\KwIn{Initial user prompt $P_0$, action space $\mathcal{A} = \{a_1, \dots, a_n\}$, maximum step number $T_{\max}$}
\KwOut{Final image $I^*$}

Initialize prompt $P \leftarrow P_0$; image $I \leftarrow \varnothing$; observation history $\mathcal{O} \leftarrow [\,]$ \\
\For{$t = 1$ \KwTo $T_{\max}$}{
    \tcp{1. Reasoning and Action Selection}
    Use the policy controller $\pi_\theta$ to determine the next action: \\
    \quad $a_t \leftarrow \pi_\theta(a \mid s_t)$ \\
    \If{$a_t =$ \texttt{STOP}}{
        break \tcp{The model decides the generation is satisfactory}
    }

    \tcp{2. Action Invocation}
    Execute the selected action ${a_t} \in \mathcal{A}$: \\
    \quad $(P_{t+1}, I_{t+1}, o_t) \leftarrow f_{a_t}(P_t, I_t, \mathcal{O}_{t-1})$ \\
    where $o_t$ is the new observation (e.g.,  quality evaluation)

    \tcp{3. State Update}
    Append $o_t$ to the observation history: $\mathcal{O}_{t} \leftarrow \mathcal{O}_{t-1} \cup \{o_t\}$
}
\Return Final image $I^* \leftarrow I$

\end{algorithm}

\begin{figure}[t]
    \centering
    \begin{tcolorbox}[
        enhanced,
        colback=green!2!white,
        colframe=green!45!black,
        title=\textbf{Prompt for Policy Controller Decision}, 
        coltitle=white,
        fonttitle=\bfseries\sffamily,
        arc=3mm,
        boxrule=0.8pt,
        left=10pt, right=10pt, top=6pt, bottom=8pt,
        fontupper=\scriptsize\sffamily,
    ]
    
    You are an \textbf{expert image generation strategist}. \\
    Your goal is to select the most suitable action for the next step in the image generation pipeline.
    \vspace{-1mm}

    \tcbline
    \vspace{-1mm}
    
    {\color{green!60!black}\textbf{\#\#\# 1. Current State}} \\
    \vspace{-2mm}

    \begin{tabular}{@{}ll@{}}
    $\bullet$ \textbf{Current Prompt:} & \texttt{"\{prompt\}"} \\
    $\bullet$ \textbf{Iteration:}      & \texttt{\{iteration + 1\} / \{max\_iteration\}} \\
    $\bullet$ \textbf{Actions so far:} & \texttt{\{action\_history\}} \\
    $\bullet$ \textbf{Prompts so far:} & \texttt{\{prompt\_history\}}
    \end{tabular}
    
    \vspace{2mm}
    {\color{green!60!black}\textbf{\#\#\# 2. Action Description}} \\
    $\triangleright$ \texttt{generate\_basic}: Generate an image directly from the current prompt. \\
    $\triangleright$ \texttt{generate\_with\_think}: Generate an image with a reasoning step that analyzes the prompt before generation. \\
    $\triangleright$ \texttt{generate\_with\_self\_assess}: Analyze the current image and refine the prompt before regenerating. \\
    $\triangleright$ \texttt{generate\_with\_self\_edit}: Generate editing instructions based on the prompt and image, then edit the image. \\
    $\triangleright$ \texttt{best\_of\_n\_selection}: Generate multiple images and select the best one to minimize error.
    
    \vspace{2mm}
    {\color{green!60!black}\textbf{\#\#\# 3. Decision Principles}} \\
    \textbf{1. If no image exists yet:} \\
    \hspace*{4mm} $\rightarrow$ Prefer \texttt{generate\_with\_think} for complex prompts. \\
    \hspace*{4mm} $\rightarrow$ Prefer \texttt{generate\_basic} for simple prompts. \\
    \textbf{2. If an image already exists but looks unsatisfactory:} \\
    \hspace*{4mm} $\rightarrow$ Prefer \texttt{generate\_with\_self\_assess} or \texttt{generate\_with\_self\_edit}. \\
    \textbf{3. If multiple generations are needed:} \\
    \hspace*{4mm} $\rightarrow$ Use \texttt{best\_of\_n\_selection}. \\
    \textbf{4. If results already look perfect:} \\
    \hspace*{4mm} $\rightarrow$ Choose \textbf{STOP}.
    
    \tcbline
    
    {\color{green!60!black}\textbf{\#\#\# Output Format}} \\
    \texttt{Reasoning: [concise explanation]} \\
    \texttt{Answer: \textbackslash boxed\{[action\_name]\}}
    
    \end{tcolorbox}
     \vspace{-4mm}
    \caption{The decision prompt for the policy controller in ImAgent.}
    \vspace{-4mm}
    \label{fig:decision_prompt}
\end{figure}

\subsection{Action Space}\label{sec:action-space}
In this section, we introduce the action space $\mathcal{A}$ of \ours. The detailed definitions and usage of each action are provided in the Appendix~\ref{sec:detailed_action_description}. The detailed prompts for each action are provided in the Appendix~\ref{sec:detailed_prompt_for_each_action}.

\vspace{-2.5mm}
\paragraph{Naive Generation/Editing.}
This action performs a one-shot image generation or editing operation directly based on the current prompt. It is typically used when the input description is simple, unambiguous, and requires no iterative refinement.

\vspace{-2.5mm}
\paragraph{Prompt Enhancement with CoT.}
This action refines the input prompt by enriching vague or underspecified user queries with additional contextual and descriptive details. Since text-to-image (T2I) models are highly sensitive to prompt wording~\cite{brade2023promptify,mo2024dynamic,liu2025one}, more elaborate and specific prompts typically lead to higher-quality visual outputs. However, most T2I backbones are trained on simple captions or surface-level image descriptions, which limits their ability to reason over complex or compositional instructions. To address this limitation, we leverage the language reasoning capability of the model’s understanding module through Chain-of-Thought (CoT) prompting. By explicitly performing intermediate reasoning and elaboration, the agent transfers structured linguistic insights from the understanding domain to the generation process, thus producing more semantically aligned and visually coherent outputs.

\vspace{-2.5mm}
\paragraph{Prompt Revision Based on the Generated/Edited Image.}
This action is triggered when the generated or edited image indicates that the current prompt is suboptimal. In this case, the UMM is prompted to analyze the discrepancy between the visual output and the intended semantics, and to self-revise the prompt accordingly. This self-correction mechanism allows the agent to iteratively refine textual descriptions based on visual feedback, thereby enhancing both semantic alignment and generation quality over successive iterations.

\vspace{-2.5mm}
\paragraph{Image Detail Refinement.}
This action targets minor visual imperfections when the input instruction is already satisfactory and the remaining issues mainly originate from the generation module. It focuses on refining local visual details such as textures, lighting, and small artifacts. Specifically, the action first generates a local editing prompt conditioned on the current image and the original user prompt, which is used internally to guide fine-grained image editing. Note that this local editing prompt is different from the user-provided prompt in image editing tasks. By enhancing fine-grained visual fidelity while preserving semantic consistency, this action improves the perceptual quality and realism of the generated results.

\vspace{-2.5mm}
\paragraph{Best-of-N Sampling.}
This action mitigates the inherent stochasticity of T2I generation. Unlike language models that typically produce stable, low-entropy outputs through supervised fine-tuning (SFT) and reinforcement learning (RL) optimization, T2I models exhibit substantial variance across different samples~\cite{tian2025unigen,verdun2025soft,li2025reflect}. To reduce this randomness, this action generates $N$ candidate images and then evaluate their visual and semantic alignment based on the given prompt. The image with the highest alignment score is selected as the final output.

\vspace{-3mm}
\paragraph{STOP.}
This action signals the end of the iterative reasoning process when the agent determines that the current image and prompt have achieved satisfactory alignment. Instead of relying on a fixed number of steps, the agent autonomously decides to terminate based on its internal evaluation of visual quality.

\vspace{-3mm}
\subsection{Policy Controller}
\vspace{-1.5mm}
\paragraph{Decision Prompt.}
We construct a structured decision prompt for the policy controller, which takes the current prompt, the generated image, and the observation history as inputs, and outputs a discrete action from the predefined action space $\mathcal{A}$. As shown in Figure~\ref{fig:decision_prompt}, the decision prompt includes the current generation prompt and the observation history (including action and prompt histories), and explicitly instructs the model to select the most suitable next action from the predefined action space.

\vspace{-2.5mm}
\paragraph{Action Parsing.}
Most UMMs are not explicitly fine-tuned for structured agentic decision-making and therefore cannot reliably output JSON-formatted actions.  To enable robust action triggering, we constrain the model to output the selected action in the form \texttt{\textbackslash boxed\{action\_name\}}. 
We then parse the action token using deterministic regular expression matching and map it to the corresponding action function.

To mitigate hallucinated, malformed, or out-of-set action outputs, we adopt a fallback mechanism: if the parsed action does not belong to the predefined action set, the system defaults to the \textit{Naive Generation/Editing} action. This design guarantees robust execution even when the policy controller produces invalid or out-of-distribution outputs. A detailed robustness analysis of the action prediction behavior of \ours\ is provided in Section~\ref{sec:robustness_analysis}.

\vspace{-3mm}
\section{Experiments}
\vspace{-1mm}
\subsection{Experimental Setting}

\vspace{-1.5mm}
\paragraph{Agentic Details.}
The maximum number of steps $T_{\max}$ is set to 5 by default, meaning that the agent will automatically select the \texttt{STOP} action if this limit is reached. 
For the \textit{Best-of-N Sampling} action, we set $N = 4$ by default. 

\vspace{-2.5mm}
\paragraph{Backbone Models.} We build our \ours on two of the most recent and powerful unified multimodal models (UMMs), Bagel~\cite{deng2025emerging} and Janus-Pro-7B~\cite{chen2025janus}. Both models possess versatile multimodal capabilities, including image understanding, image generation, which collectively form the foundation of our framework. For image generation, we leverage both Bagel and Janus-Pro-7B. However, since Janus-Pro-7B does not support image editing, we employ Bagel exclusively for the image editing tasks.

\vspace{-2.5mm}
\paragraph{Benchmarks and Baselines.}
We evaluate \ours on multiple benchmarks, including 3 benchmarks on image generation and 4 benchmarks for image editing tasks. For image generation, we use R2I-Bench~\cite{chen2025r2i}, WISE~\cite{niu2025wise}, and T2I-ReasonBench~\cite{sun2025t2i}. For image editing, we adopt GEdit-Bench~\cite{liu2504step1x}, RISEBench~\cite{zhao2025envisioning}, KRIS-Bench~\cite{wu2025kris}, and ImgEdit-Bench~\cite{ye2025imgedit}. We utilize various challenging models as baselines for image generation and image editing comparison. Additionally, we compare \ours with the backbone UMM and a prompt \textit{Self-rewrite} baseline implemented using the same backbone model. The details are provided in Appendix~\ref{sec:Detailed_Benchmarks_Baselines}.

\begin{table*}[t]
  \centering
  \small

    \caption{Experimental results on \textbf{R2I-Bench}~\cite{chen2025r2i}. \textit{Comm.} and \textit{Comp.} denote the Commonsense and Compositional categories, respectively. Due to the inaccessibility of parts of the Mathematical and Concept-Mixing categories, these two categories are omitted. Vanilla represents the backbone model, while \ours denotes our agent built upon this model. The best performance is highlighted in \textbf{bold}.}
    \vspace{-2.5mm}

  \setlength{\tabcolsep}{3pt}
    \resizebox{\textwidth}{!}{
  
    \begin{tabular}{lccccccc}
    \hline
    
    \hline
      Types & Model & \multicolumn{1}{l}{Comm.} & \multicolumn{1}{l}{Comp.} & \multicolumn{1}{l}{Logical} & \multicolumn{1}{l}{Numerical} & \multicolumn{1}{l}{Causal} & \multicolumn{1}{l}{Overall}  \bigstrut[t]\\
    \hline
    & SD3-medium~\cite{rombach2022high} & 0.54 & 0.64 & 0.55 & 0.50  & 0.18 & 0.53 \\
    & Sana-1.5~\cite{xie2025sana} & 0.49 & 0.67  & 0.49 & 0.48 &  0.21 & 0.49 \\
    & Lumina-T2I~\cite{qin2025lumina} &  0.38 & 0.49  & 0.38 & 0.45 &  0.18 & 0.39 \\
    & Omnigen~\cite{xiao2025omnigen} & 0.43 & 0.60  & 0.51 & 0.47 & 0.34 & 0.48 \\
    & LLM4GEN$_{SD1.5}$~\cite{liu2025llm4gen} & 0.55 & 0.48  & 0.55 & 0.39 & 0.45  & 0.51 \\
    & ELLA$_{SD1.5}$~\cite{hu2024ella} & 0.40 & 0.44  & 0.40 & 0.32 & 0.29 & 0.39  \\ 
    & LlamaGen~\cite{sun2024autoregressive} & 0.38 & 0.39  & 0.38 & 0.35 & 0.12 & 0.36 \\
    & DALL-E-3~\cite{ma2025learning} & 0.78 & 0.76  & 0.69 & 0.69  & 0.64 & 0.73 \\
    \multirow{-9}{*}{\textit{Gen}}  & gpt-image-1~\cite{hurst2024gpt} &  0.83 & 0.87  & 0.81 & 0.88  & 0.71 & 0.83 \\

    \hline
    & EMU3~\cite{wang2024emu3} & 0.44 & 0.59 & 0.55 & 0.61 &  0.41 & 0.52 \\
    & Show-o~\cite{xie2024show} & 0.42 & 0.59  & 0.42 & 0.57 &  0.30 & 0.46 \\
    \multirow{-3}{*}{\textit{Unified}} & Lumina-Image 2.0~\cite{qin2025lumina} &  0.49 & 0.65  & 0.56 & 0.43 &  0.40  & 0.52 \\
    
    \hline
    \multirow{3}{*}{\textit{Bagel}}
    & Vanilla~\cite{deng2025emerging} & 0.47 & 0.65  & 0.53 & 0.68 &  0.40 & 0.54 \\
    & Self-Rewrite~\cite{deng2025emerging} & 0.54 & 0.67 & 0.54 & 0.69 & 0.47 & 0.58 \\
    & \ours (Ours) &  \textbf{0.58} & \textbf{0.68} & \textbf{0.62} & \textbf{0.71} & \textbf{0.53} & \textbf{0.62} \\

    \hline

    \multirow{3}{*}{\textit{Janus-Pro-7B}} & Vanilla~\cite{chen2025janus} &  0.45 & \textbf{0.60}  & 0.46 & 0.46 & 0.36 & 0.47 \\ 
    & Self-Rewrite~\cite{chen2025janus} & 0.40 & 0.48 & 0.35 & 0.39 & 0.28 & 0.38 \\
    & \ours (Ours) & \textbf{0.47} & 0.58   & \textbf{0.52} & \textbf{0.54} & \textbf{0.44} & \textbf{0.51} \\

    \hline

    \hline
    \end{tabular}}

    \vspace{-4mm}
   
  \label{tab:R2I}
\end{table*}

\begin{figure}[t]
    \centering
    \includegraphics[width=0.99\linewidth]{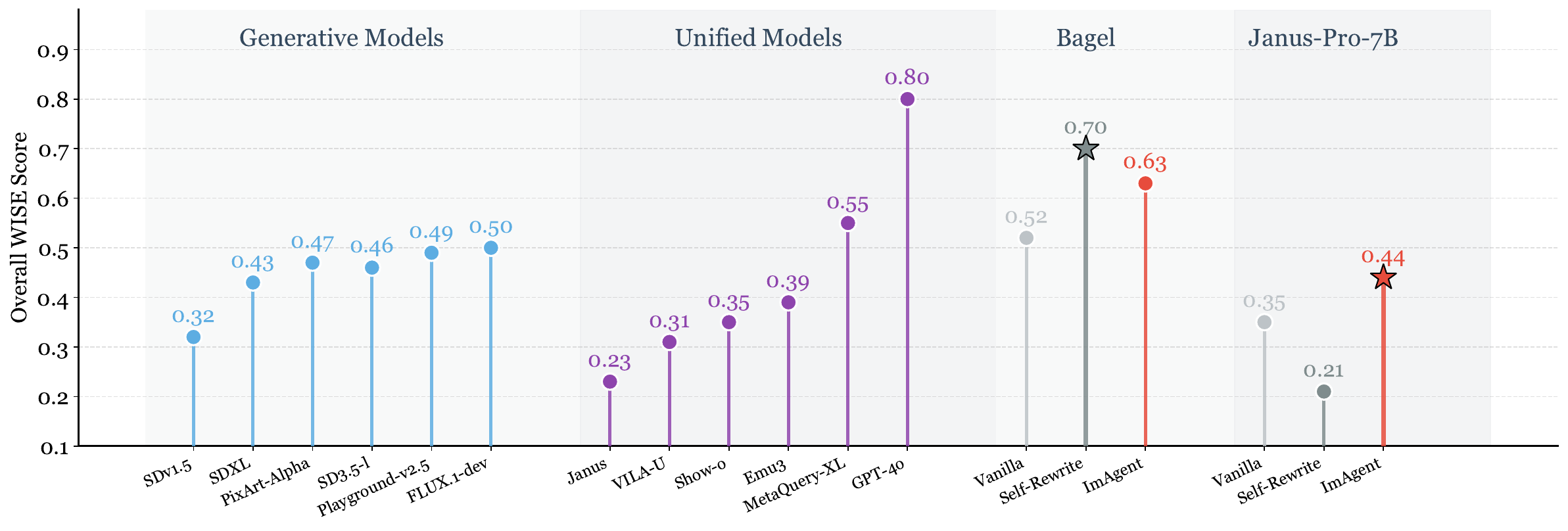}
    \vspace{-2mm}
    \caption{Experimental results on \textbf{WISE}~\cite{niu2025wise}. Vanilla represents the backbone model, while \ours denotes our agent built upon this model.}
    \vspace{-2mm}
    \label{fig:wise}
\end{figure}

\begin{table*}[t]
  \centering
  \small

      \caption{Experimental results on \textbf{T2I-ReasonBench}~\cite{sun2025t2i}. Vanilla represents the backbone model, while \ours denotes our agent built upon this model.}

        \vspace{-3mm}

  \setlength{\tabcolsep}{3pt} 
\resizebox{\textwidth}{!}{ 
    \begin{tabular}{lccccccccccc}
    \hline
    
    \hline
      \multirow{2}{*}{Types} & \multirow{2}{*}{Model}  & \multicolumn{2}{c}{Idiom} & \multicolumn{2}{c}{Textual} & \multicolumn{2}{c}{Entity}  & \multicolumn{2}{c}{Scientific} & \multicolumn{2}{c}{Overall} \\
      \cline{3-12}
      
      &  & Acc.& Qual. & Acc.& Qual. & Acc.& Qual. & Acc.& Qual. & Acc.& Qual. \bigstrut[t]\\
    \hline

    
    & FLUX.1-dev~\cite{flux2024} & 39.1 & 83.4 & 56.9 & 76.5 & 45.1 & 90.6 & 46.7 & 80.9 & 47.0& 82.8 \\
    & FLUX.1-schnell~\cite{flux2024} & 40.9 & 83.1& 65.1& 74.5& 44.8& 91.5 &50.7& 83.0& 50.4& 83.0 \\

    & playground-v2.5~\cite{li2024playground} & 43.9& 87.8& 38.5& 72.1& 48.4& 92.4& 50.8& 83.3& 45.4 &83.9 \\

    & SD-3-Medium~\cite{rombach2022high} & 35.9& 81.4& 60.9& 71.3& 42.4& 90.1& 50.9& 81.7 &47.5& 81.1 \\

    & SD-3.5-Medium~\cite{esser2024scaling} & 34.4 &80.6& 58.0& 70.1& 44.8& 92.1& 49.9& 83.0& 46.8 &81.4 \\

    & SD-3.5-Large~\cite{esser2024scaling} & 35.6 & 85.3& 62.2& 75.4& 46.6& 92.6& 52.9 &84.5 &49.3& 84.4 \\

    \multirow{-7}{*}{\textit{Gen}}  & gpt-image-1~\cite{hurst2024gpt} & 75.7 & 94.5 & 86.9 & 97.6 & 77.5 & 96.6 & 74.7 & 94.3 & 78.7&  95.8 \\

    \hline

    \multirow{4}{*}{\textit{Unified}}
    & Emu3~\cite{wang2024emu3} & 33.1 &82.9& 33.7& 68.7& 33.8& 85.2& 40.1& 77.0& 35.2 &78.5 \\

    & show-o~\cite{xie2024show} & 33.1 &82.5& 35.3& 80.3& 34.9 &87.4& 41.6& 76.6& 36.2 &81.7 \\

    &  GoT~\cite{fang2025got} & 29.7 &76.4& 30.6 &70.7& 31.0& 86.2& 36.8& 76.3 &32.0& 77.4 \\

    & Gemini-2.0~\cite{google2025_gemini2_image} & 52.4 & 87.8 & 73.0 & 83.3  & 67.0 & 94.3 & 66.7 & 89.3  & 64.8 & 88.7 \\

    \hline

    \multirow{3}{*}{\textit{Bagel}}
    & Vanilla~\cite{deng2025emerging} & 30.2 &  85.7 &  36.6 & 68.4 & 45.0 & 94.7 & 54.4& 87.5 & 41.6 & 84.1 \\
    & Self-Rewrite~\cite{deng2025emerging} & \textbf{44.6} & 84.3& 44.0& 73.7& 52.4& 91.6& 57.7& 88.3& 49.7& 84.5 \\

    & \ours (Ours) & 37.7  & \textbf{90.0}  & \textbf{54.2}  &  \textbf{79.1} & \textbf{52.6} & \textbf{96.6}& \textbf{61.2} & \textbf{90.3} & \textbf{51.4}  & \textbf{89.0}\\

    \hline

    \multirow{3}{*}{\textit{Janus-Pro-7B}}
    & Vanilla~\cite{chen2025janus} &  25.5 &  78.0  &  \textbf{37.2} & \textbf{70.9} & 38.5  & 87.6  & 44.9 & 77.8 & 36.5 & 78.6 \\
    
    & Self-Rewrite~\cite{chen2025janus} & 11.9 & 74.6 & 9.2 & 49.3 & 12.0 & 74.3 & 14.0& 59.3 &11.1 & 63.3 \\

     & \ours (Ours) & \textbf{27.9}  & \textbf{86.0}  & 35.3  & 68.8 & \textbf{40.7} & \textbf{89.7}  & \textbf{51.2} & \textbf{84.2} & \textbf{38.8} & \textbf{82.2}\\
    
    \hline

    \hline
    \end{tabular}}

    \vspace{-4mm}

  \label{tab:t2i-reasoning}
\end{table*}

\begin{table*}[t]
  \centering
  \small

    \caption{
    Experimental results on \textbf{RISEBench}~\cite{zhao2025envisioning}. Vanilla represents the backbone model, while \ours denotes our agent built upon this model. The best performance between vanilla and \ours is highlighted in \textbf{bold}.
  }
  \vspace{-2.5mm}

  \setlength{\tabcolsep}{4pt}
\resizebox{0.8\textwidth}{!}{

    \begin{tabular}{lcccccc}
    \hline
    
    \hline

      Types & Model & \multicolumn{1}{l}{Temporal} & \multicolumn{1}{l}{Causal} & \multicolumn{1}{l}{Spatial} & \multicolumn{1}{l}{Logical} & \multicolumn{1}{l}{Overall} \bigstrut[t]\\
    \hline
    \multirow{3}{*}{\textit{Private}} 
    & Gemini-2.0-Flash-pre~\cite{team2023gemini} & 10.6 & 13.3 & 11.0 & 2.3 & 9.4\\
    & Gemini-2.0~\cite{google2025_gemini2_image} & 8.2 & 15.5& 23.0 & 4.7 & 13.3 \\
     & GPT-4o~\cite{openai2025gpt4o_image} & 34.1 & 32.2 & 37.0 & 10.6 & 28.9 \\
    \hline
    \multirow{7}{*}{\textit{Open}}
    & EMU2~\cite{sun2024generative} & 1.2 & 1.1 & 0.0 & 0.0 & 0.5 \\
    & OmniGen~\cite{xiao2025omnigen} & 1.2 & 1.0 & 0.0 & 1.2 & 0.8 \\
    & Step1X-Edit~\cite{liu2025step1x} & 0.0 & 2.2 & 2.0 & 3.5 & 1.9 \\
    & Qwen-Image-Edit~\cite{wu2025qwenimagetechnicalreport} & 4.7 & 10.0 &17.0 & 2.4 & 8.9 \\
    & FLUX.1-Kontext-Dev~\cite{labs2025flux1kontextflowmatching} & 2.3 & 5.5 & 13.0 & 1.2 & 5.8 \\
    & Ovis-U1~\cite{wang2025ovisu1} & 1.2 & 3.3 & 4.0 & 2.4 & 2.8 \\
    & Seedream-4.0~\cite{seedream2509seedream} & 12.9 & 12.2 & 11.0 & 7.1 & 10.8 \\
    \hline
    \multirow{3}{*}{\textit{Bagel}}
    & Vanilla~\cite{deng2025emerging} & 2.4 & 5.6 & 14.0 & 1.2 & 6.1 \\
    & Self-Rewrite~\cite{deng2025emerging} & 5.9 & \textbf{17.8} & \textbf{21.0} & 1.2 & 11.9 \\
   & \ours (Ours) &  \textbf{17.6} & 15.6 & 16.0 & \textbf{2.4} & \textbf{13.1}  \\

    \hline

    \hline
    \end{tabular}}

  \label{tab:rise_edit}
\end{table*}

\begin{table*}[t]
\centering

\caption{Experimental results on \textbf{ImgEdit-Bench}~\cite{ye2025imgedit}.}
  \vspace{-2.5mm}

\setlength{\tabcolsep}{3pt}
\resizebox{\textwidth}{!}{

\begin{tabular}{lcccccccccc|c}

\hline

\hline
 \textbf{Types} & \textbf{Models} & \textbf{Add} & \textbf{Adjust} & \textbf{Extract} & \textbf{Replace} & \textbf{Remove} & \textbf{Background} & \textbf{Style} & \textbf{Hybrid} & \textbf{Action} & \textbf{Overall} \\

 \hline

 \multirow{1}{*}{\textit{Private}} &  GPT-4o~\cite{openai2025gpt4o_image} & 4.61 & 4.33 & 2.90 & 4.35 & 3.66 & 4.57 & 4.93 & 3.96 & 4.89 & 4.20 \\

 \hline

 \multirow{6}{*}{\textit{Open}}
&  MagicBrush~\cite{zhang2023magicbrush} & 2.84 & 1.58 & 1.51 & 1.97 & 1.58 & 1.75 & 2.38 & 1.62 & 1.22 & 1.83 \\
&  Instruct-P2P~\cite{brooks2023instructpix2pix} & 2.45 & 1.83 & 1.44 & 2.01 & 1.50 & 1.44 & 3.55 & 1.20 & 1.46 & 1.88 \\
&  AnyEdit~\cite{yu2025anyedit} & 3.18 & 2.95 & 1.88 & 2.47 & 2.23 & 2.24 & 2.85 & 1.56 & 2.65 & 2.45 \\
& UltraEdit~\cite{zhao2024ultraedit} & 3.44 & 2.81 & 2.13 & 2.96 & 1.45 & 2.83 & 3.76 & 1.91 & 2.98 & 2.70 \\
& Step1X-Edit~\cite{liu2025step1x} & 3.88 & 3.14 & 1.76 & 3.40 & 2.41 & 3.16 & 4.63 & 2.64 & 2.52 & 3.06 \\
& UniWorld-V1~\cite{lin2025uniworld} & 3.82 & 3.64 & 2.27 & 3.47 & 3.24 & 2.99 & 4.21 & 2.96 & 2.74 & 3.26 \\

\hline

\multirow{3}{*}{\textit{Bagel}} 
& Vanilla~\cite{deng2025emerging} & \textbf{3.89} & 3.68 & 1.33 & \textbf{3.78} & \textbf{3.07} & 2.07 & \textbf{4.30} & 2.47 & 4.33 & 2.89 \\

& Self-Rewrite~\cite{deng2025emerging} & 3.22 & 3.25 & \textbf{2.31} & 3.55 & 2.13 & 3.41 & 3.97 & 2.61 & \textbf{4.56} & 3.00 \\

& ~\ours (Ours) & 3.20 & \textbf{3.89} & 1.77 & 3.13 & 2.71 & \textbf{3.50} & 4.21 & \textbf{2.89} & 4.43 & \textbf{3.15} \\

\hline

\hline

\end{tabular}}

\vspace{-4mm}
\label{tab:image_edit_comparison}
\end{table*}

\begin{table*}[t]
  \centering

    \caption{
    Experimental results on \textbf{GEdit-Bench}~\cite{liu2504step1x}, covering both English (EN) and Chinese (CN) settings. 
    “--” denotes that the model does not support Chinese.}
    \vspace{-2.5mm}

  \setlength{\tabcolsep}{5pt}
\resizebox{0.7\textwidth}{!}{

  \begin{tabular}{ll|ccc|ccc}
    \hline

    \hline
    \multirow{2}{*}{Types} 
    & \multirow{2}{*}{Models} & \multicolumn{3}{c|}{EN Setting} & \multicolumn{3}{c}{CN Setting} \\ 
    \cline{3-8}
    &  & G\_SC & G\_PQ & G\_O & G\_SC & G\_PQ & G\_O \\
    \hline

    \multirow{3}{*}{\textit{Private}} 
      & Gemini-2.0~\cite{google2025_gemini2_image}  & 6.73 & 6.61 & 6.32 & 5.43 & 6.77 & 5.36 \\
      & Doubao~\cite{shi2411seededit}  & 6.92 & 7.19 & 6.75 & 6.98 & 7.27 & 6.77 \\ 
      & GPT-4o~\cite{openai2025gpt4o_image}  &  7.85 & 7.62 & 7.53 & 7.67 & 7.56 & 7.30 \\
    \hline

    \multirow{5}{*}{\textit{Open}} 
      & Instruct-Pix2Pix~\cite{brooks2023instructpix2pix} & 3.58 & 5.50 & 3.68 & - & - & - \\
      & MagicBrush~\cite{zhang2023magicbrush}       & 4.68 & 5.66 & 4.52 & - & - & - \\
      & AnyEdit~\cite{yu2025anyedit}          & 3.18 & 5.82 & 3.21 & - & - & - \\
      & OmniGen~\cite{xiao2025omnigen}          & 5.96 & 5.89 & 5.06 & - & - & - \\
      & Step1X-Edit~\cite{liu2025step1x}      & 7.09 & 6.76 & 6.70 & 7.20 & 6.87 & 6.86 \\
    \hline

    \multirow{3}{*}{\textit{Bagel}} 
      & Vanilla~\cite{deng2025emerging}  & 7.36 & \textbf{6.83} & 6.52  & 7.34 & \textbf{6.85} & 6.50  \\
      & Self-Rewrite~\cite{deng2025emerging} & 7.66  & 6.57 & 6.61 &  7.50 &  6.60  & 6.58 \\
      & \ours (Ours) & \textbf{7.85} & 6.61 & \textbf{6.88}& \textbf{7.92} & 6.53 & \textbf{6.84} \\
    \hline

    \hline
  \end{tabular}}

  \label{tab:gedit}
\end{table*}




\vspace{-2mm}
\subsection{Main Results}

\vspace{-1mm}
\paragraph{Image Generation.}
As shown in Figure~\ref{fig:wise}, \ours achieves outstanding performance on the WISE benchmark. Compared to their respective backbone models, \ours built upon Bagel and Janus-Pro-7B consistently surpasses the vanilla counterparts, with performance improvements of 21.2\% and 25.7\%, respectively, demonstrating the effectiveness of our approach in enhancing image generation quality through coordinated reasoning and refinement. Furthermore, the vanilla Janus-Pro-7B underperforms compared to SDXL and EMU3 and performs on par with Show-o. In contrast, \ours constructed on Janus-Pro-7B surpasses all of them, highlighting its strong test-time scaling capability and generalization potential. Furthermore, on the reasoning benchmarks, \ours consistently delivers strong results. As shown in Table~\ref{tab:R2I} and Table~\ref{tab:t2i-reasoning}, \ours outperforms the vanilla models across both backbone architectures. For instance, \ours achieves improvements of 14.8\% and 7.5\% across both backbones on R2I-Bench, respectively. Moreover, \ours built upon Bagel surpasses the commercial model Gemini-2.0 in overall quality on the T2I-ReasonBench, further demonstrating its effectiveness in reasoning-based image generation. 

Notably, \ours constructed on Bagel underperforms the Self-Rewrite baseline on WISE. To further investigate this phenomenon, we perform an additional control experiment by rewriting the prompts using GPT-4o with the same system prompt as used in the Bagel Self-Rewrite implementation. Surprisingly, this GPT-4o-based rewrite achieves a score of 0.64, which is significantly lower than the Self-Rewrite performance. This suggests that the strong performance of Self-Rewrite on this dataset may benefit from dataset-specific familiarity. Additionally, Self-Rewrite underperforms Vanilla on Janus-Pro-7B, whereas \ours constructed on Janus-Pro-7B consistently outperforms Vanilla, using the same prompt as in the \textit{Prompt Enhancement with CoT} action. This observation suggests that Janus-Pro-7B is less effective in text-only prompt rewriting, while \ours mitigates this limitation by leveraging test-time agentic control and scalable decision-making.


\vspace{-1mm}
\paragraph{Image Editing.}
As shown in Table~\ref{tab:rise_edit}, vanilla Bagel achieves only 6.1 on RISEBench, underperforming compared to Qwen-Image-Edit and Seedream-4.0. In contrast, \ours achieves a 114.8\% improvement over vanilla and surpasses both Qwen-Image-Edit and Seedream-4.0. Remarkably, \ours performs on par with Gemini-2.0, demonstrating its effectiveness and strong test-time scaling capability. This indicates that open-source models enhanced with \ours can achieve competitive performance with commercial models.

Furthermore, \ours performs well on ImgEdit-Bench (shown in Table~\ref{tab:image_edit_comparison}) as well, with vanilla achieving 2.89 and underperforming compared to Step1X-Edit, while \ours reaches 3.15, outperforming Step1X-Edit and achieving the best performance among open-source models. This further demonstrates the effectiveness of \ours in test-time scaling. Moreover, as shown in Table~\ref{tab:gedit} \ours outperforms vanilla Bagel across both English and Chinese settings in GEdit-Bench, achieving improvements of 5.5\% and 5.2\%, respectively. Detailed results are provided in the Appendix.


\begin{figure}[t]
    \centering
    \includegraphics[width=0.98\linewidth]{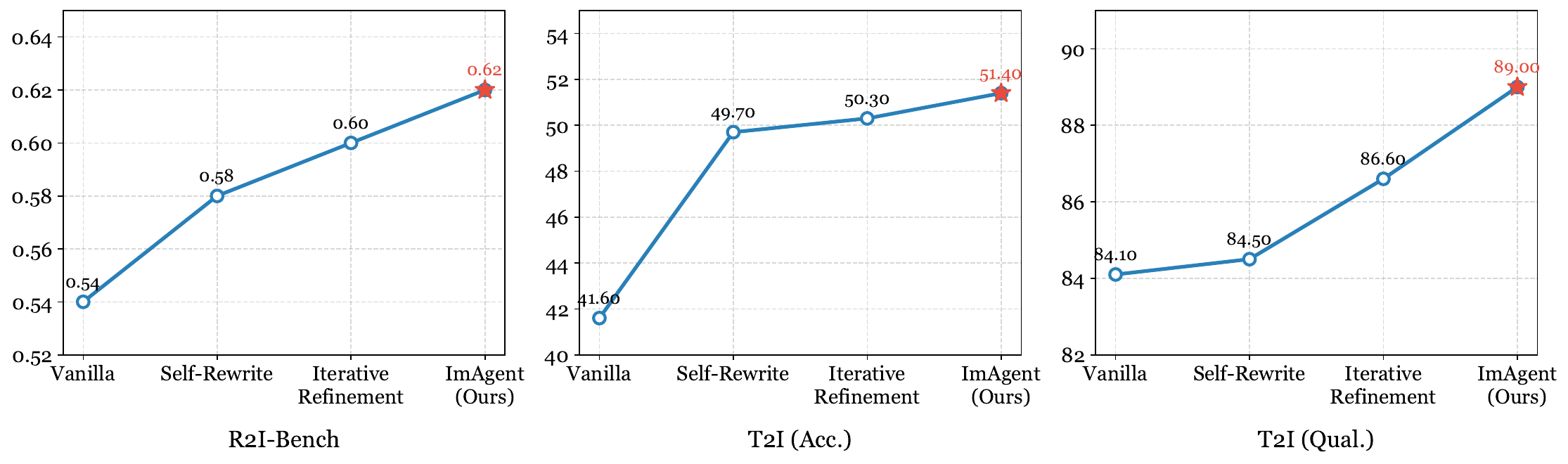}
    \vspace{-2.5mm}
    \caption{Comparison between iterative prompt refinement and other baselines on Bagel across R2I-Bench and T2I-ReasonBench.}
    \vspace{-4mm}
    \label{fig:iterative-refinement}
\end{figure}

\vspace{-3mm}
\subsection{Robustness Analysis}\label{sec:robustness_analysis}
Beyond the strong quantitative results across various benchmarks, robustness is critical for the reliable deployment and generalization of training-free architectures. We therefore conduct a robustness analysis of \ours using three metrics: Parse Success Rate (PSR), Action Uniqueness Rate (AUR), and Action Validity Rate (AVR). PSR measures whether the model successfully parses an action, AUR evaluates whether the model outputs a single valid action rather than hallucinating multiple actions, and AVR measures the probability that the predicted action belongs to the predefined action set. The detailed definitions of these metrics are available in Appendix.

Table~\ref{tab:robustness_analysis} reports the robustness results under different backbones and datasets. 
We observe near-perfect PSR, AUR, and AVR scores across all evaluated settings, indicating that the policy controller reliably follows the structured decision prompt and produces valid, non-ambiguous actions. 
In particular, the high PSR demonstrates that deterministic parsing via \texttt{\textbackslash boxed\{\}} is highly effective, while the near-100\% AUR and AVR confirm that the model rarely hallucinates multiple or invalid actions. 
These results suggest that structured decision prompting and deterministic parsing can effectively stabilize policy behavior even in a fully training-free setting.

\begin{figure}[t]
    \centering
    \begin{tcolorbox}[
        enhanced,
        colback=green!2!white,
        colframe=green!45!black,
        title=\textbf{Example: Policy Controller Output},
        coltitle=white,
        fonttitle=\bfseries\sffamily,
        arc=3mm,
        boxrule=0.8pt,
        left=10pt, right=10pt, top=8pt, bottom=8pt,
        fontupper=\footnotesize\sffamily, 
    ]
    
    \textbf{Reasoning:} The current image does not match the prompt as it requires a bread with a greenish or blackish mold, but the image provided does not clearly show the mold. The current image may be unsatisfactory or incomplete. Therefore, the best course of action is to use \textbackslash boxed\{prompt$\_$refinement\} to analyze the current image and refine the prompt before regenerating.
    
    \vspace{1mm}
    \textbf{Extracted Answer:} \texttt{\{prompt\_refinement\}}

    \end{tcolorbox}
    \vspace{-4mm}
    \caption{An example of the policy controller making a decision.}
    \label{fig:policy-decision}
    \vspace{-1mm}
\end{figure}

\begin{figure}[t]
    \centering
    \includegraphics[width=0.98\linewidth]{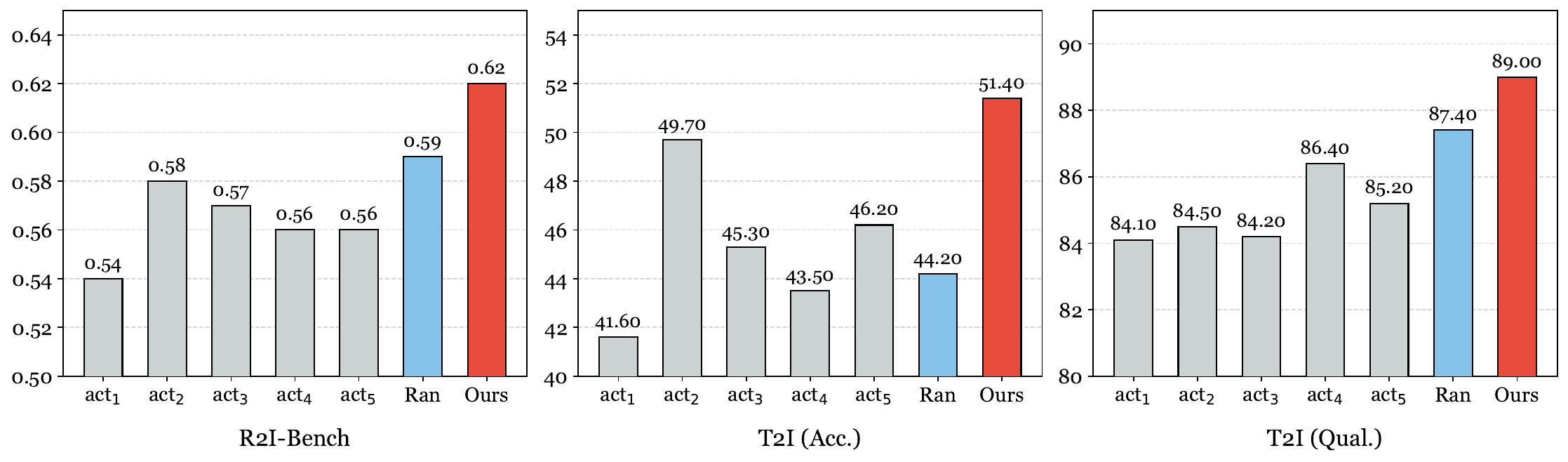}
    \vspace{-4mm}
    \caption{Comparison of different action selection policies on Bagel. action$_1$--action$_5$ represent the five actions (Section~\ref{sec:action-space}) in order.}
    \vspace{-4mm}
    \label{fig:ablation2}
\end{figure}

\begin{table}[t]
\centering

\caption{Robustness analysis of the policy controller. We report Parse Success Rate (PSR), Action Uniqueness Rate (AUR), and Action Validity Rate (AVR) on WISE and RISEBench under Bagel and Janus backbones.}
\vspace{-3.5mm}

\setlength{\tabcolsep}{5pt}
\resizebox{0.56\textwidth}{!}{

\begin{tabular}{cccc}
\hline

\hline
Dataset & PSR (\%) & AUR (\%) & AVR (\%) \\
\hline
WISE$_{Bagel}$ & 100\% & 100\% & 100\% \\
WISE$_{Janus}$ &  99\% & 100\% & 100\% \\
RISEBench$_{Bagel}$ & 100\% & 100\% & 100\% \\
\hline

\hline
\end{tabular}}

\vspace{-6mm}
\label{tab:robustness_analysis}
\end{table}

\vspace{-2mm}
\subsection{Agentic Behavior Analysis}
We compare the agentic behaviors of \ours constructed on different UMMs by analyzing their decision trajectories. We derive two key observations. \\
\textbf{Observation 1:} The average decision length differs across backbone models. Specifically, \ours built on Bagel consumes more computational budget than Janus-Pro-7B. Under the image generation task, \ours on Bagel executes an average of 4.73 iterations, whereas \ours on Janus-Pro-7B requires only 3.98 iterations on average. This result indicates that, even with identical workflows and prompts, different unified backbones exhibit distinct agentic behaviors and computational resource utilization patterns. We hypothesize that Bagel tends to perform more conservative or iterative refinement, while Janus-Pro-7B exhibits more decisive behavior, leading to fewer interaction steps.\\
\textbf{Observation 2:} Both backbone models exhibit similar action selection trends. In particular, they tend to invoke the \textit{Best-of-N Sampling} action at later stages of the interaction, unless computational constraints prevent its execution, as the policy controller recognizes that this action incurs higher computational cost. Moreover, \ours invokes the \textit{Image Detail Refinement} action less frequently than other actions. This behavior is intuitive, as this action requires the model to first generate a localized editing prompt conditioned on both the generated image and the user prompt, and then perform image editing, which introduces additional complexity and uncertainty. This suggests that the \ours implicitly prefers lower-cost and lower-risk actions, resembling a rational cost–benefit trade-off strategy.

\vspace{-4mm}
\subsection{Test-Time Scalability Analysis}
\vspace{-2mm}
To demonstrate the test-time scalability of \ours, we compare it with a strong training-free iterative refinement baseline under the same inference budget. Specifically, we implement an iterative prompt self-refinement strategy, where the UMM repeatedly rewrites the prompt based on the generated image and regenerates the image for $T_{max}$ steps, matching the number of model calls used by \ours. As shown in Figure~\ref{fig:iterative-refinement}, \ours consistently outperforms iterative refinement under the same compute budget. These results indicate that adaptive policy-driven test-time scaling is more effective than naive iterative refinement.







\vspace{-3mm}
\subsection{Effectiveness of the Policy Controller of \ours}
To evaluate the effectiveness of \ours in coordinately selecting and executing appropriate actions, we compare its performance with that of single-turn actions and a random action selection policy, where the $T_{\max}$ for random selection is set the same as in \ours. As shown in Figure~\ref{fig:ablation2}, experiments are conducted on R2I-Bench and T2I-ReasonBench. Taking R2I-Bench as an example, all single-turn actions perform well compared to the naive generation action, which serves as the lower bound (0.54). Moreover, the random action selection policy achieves a score of 0.59, higher than all single-turn actions, indicating that naively constructing an agent is beneficial. Finally, \ours outperforms the random policy with a score of 0.62, demonstrating that it effectively and efficiently boosts test-time scaling in image generation.

\vspace{-3mm}
\subsection{Qualitative Analysis}\label{sec:quality}

As shown in Figure~\ref{fig:example}, we provide several qualitative examples, including two for image generation and two for image editing, each executed over five iterative steps. Overall, we observe that the image quality improves progressively at each step, demonstrating that \ours gradually refines the output. 

Specifically, consider a case where the user intends to generate an image based on the prompt ``moldy bread.'' Here, \ours dynamically selects a sequence of five actions to generate the final image. The complete decision history for this case is provided in Appendix~\ref{sec:policy_decision}. To illustrate the policy controller’s reasoning, Figure~\ref{fig:policy-decision} visualizes how the controller selects actions at the specific step. For instance, it chooses the \texttt{prompt$\_$refinement} action to enrich the textual prompt, thereby improving semantic clarity for subsequent generation steps. Additional examples of policy decisions are presented in Appendix~\ref{sec:policy_decision}.

\vspace{-3mm}
\section{Conclusion}
In this paper, we introduced \ours, a training-free unified multimodal agent designed to enhance the image generation through efficient test-time scaling. By integrating multiple generation actions within a single framework and coordinating them via a policy controller, \ours dynamically selects and executes the most appropriate action for a given case. Extensive experiments on image generation and image editing benchmarks demonstrate that \ours consistently improves over the backbone models and outperforms strong baseline methods, including scenarios where the backbone fails. These results highlight the potential of unified multimodal agents for adaptive, efficient, and high-quality image generation without relying on additional external models.


%
%
\bibliographystyle{splncs04}
\bibliography{main}

\appendix
\newpage
\section{Algorithm for Image Editing} As shown in Algorithm~\ref{alg:image_edit_agent}, we present the detailed algorithm of \ours for image editing task.

\begin{algorithm}[h]
\caption{\ours for Image Editing.}
\label{alg:image_edit_agent}
\KwIn{Initial image $I_0$, user instruction $P_0$, action space $\mathcal{A} = \{a_1, \dots, a_n\}$, maximum iteration number $T_{\max}$}
\KwOut{Final edited image $I^*$}

Initialize prompt $P \leftarrow P_0$; image $I \leftarrow I_0$; observation history $\mathcal{O} \leftarrow [\,]$ \\
\For{$t = 1$ \KwTo $T_{\max}$}{
    \tcp{1. Reasoning and Action Selection}
    Use the policy model $\pi_\theta$ to determine the next action: \\
    \quad $a_t \leftarrow \pi_\theta(\alpha \mid s_t)$ \\
    \If{$a_t =$ \texttt{STOP}}{
        break \tcp{The model decides the editing is sufficient}
    }

    \tcp{2. Action Invocation}
    Execute the selected function ${a_t} \in \mathcal{A}$: \\
    \quad $(P_{t+1}, I_{t+1}, o_t) \leftarrow f_{a_t}(P_{t}, I_{t}, \mathcal{O}_{t-1})$ \\
    where $o_t$ is the new observation

    \tcp{3. State Update}
    Append $o_t$ to the observation history: $\mathcal{O}_t \leftarrow \mathcal{O}_{t-1} \cup \{o_t\}$
}
\Return final edited image $I^* \leftarrow I$
\end{algorithm}

\section{Detailed Action Description}\label{sec:detailed_action_description}
\paragraph{Naive Generation/Editing.}
This action performs a direct, one-shot invocation of the underlying text-to-image or image-editing backbone without any intermediate reasoning or prompt restructuring. Given the current prompt–image pair $(P_t, I_t)$, the backbone generator $g(\cdot)$ produces the next image deterministically as:
\begin{equation}
I_{t+1} = g(P_t, I_t).
\end{equation}
This action is effective when the user intent is already explicit, the visual goal is straightforward, or iterative refinement offers limited benefit.

\paragraph{Prompt Enhancement with CoT.}
This action enriches the input prompt using the model’s multimodal reasoning capability. Unlike naive generation, it introduces an intermediate reasoning stage, where the model generates a chain-of-thought (CoT) enhanced prompt $E_t$ that clarifies ambiguous instructions, decomposes complex compositions, and fills missing semantic details:
\begin{equation}
E_t = \mathrm{CoT}(P_t).
\end{equation}
The intuition is that while T2I models excel at visual synthesis, they often struggle with compositional or underspecified instructions due to their training bias toward short captions. By transferring the model’s linguistic reasoning ability to the generative pathway, CoT enhancement produces prompts that are more explicit, structured, and visually actionable.
Finally, the updated prompt is fed into the generator:
\begin{equation}
I_{t+1} = g(E_t, I_t).
\end{equation}
This action is particularly beneficial when the input contains complex spatial relationships, multi-object instructions, stylistic requirements, or vague conceptual queries. By exposing the implicit reasoning steps, the model produces outputs that are more semantically aligned, compositionally consistent, and visually faithful to user intent.

\paragraph{Prompt Revision Based on the Generated/Edited Image.}

This action is invoked when the current visual output reveals a mismatch between the generated image and the intended semantics. In such cases, the UMM performs discrepancy-aware prompt revision. Specifically, the model first extracts a semantic summary from the generated image:
\begin{equation}
\hat{S}_t = \mathrm{VisionAnalyze}(I_t),
\end{equation}
and evaluates its alignment with the intended description encoded in the prompt:
\begin{equation}
\delta_t = \mathrm{Diff}(\hat{S}_t, P_t),
\end{equation}
where $\delta_t$ captures the semantic gap (e.g., missing attributes, incorrect object relations). The revised prompt $R_t$ is then formulated as:
\begin{equation}
R_{t} = \mathrm{Revise}(P_t, \delta_t).
\end{equation}
This action enables the agent to iteratively correct textual instructions based on visual feedback.
This self-correcting loop strengthens consistency between linguistic intent and visual realization, leading to progressively refined generations across steps.

\paragraph{Image Detail Refinement.}

This action aims to enhance the fidelity of local image details while preserving the original semantic content. 
When the prompt is already accurate but the generated image contains minor imperfections, such as texture noise, inconsistent lighting, or small structural artifacts, the agent performs detail-aware refinement to improve visual quality.

Formally, given the current image $I_t$ and prompt $P_t$, the UMM first leverages its multimodal reasoning capability to derive an image editing prompt $P_{edit}$ that targets the identified imperfections. The model then applies its image editing ability to refine the image accordingly:
\begin{equation}
\left\{
\begin{aligned}
& P_{edit} = R(P_t, I_t) \\
& I_{t+1} = r(I_t, P_{edit})
\end{aligned}
\right.
\end{equation}

where $R(\cdot)$ denotes the reasoning process that produces the editing instruction, and $r(\cdot)$ represents the image editing operation. This action improves perceptual quality without altering the overall scene composition, making it particularly effective for polishing near-final results.

\paragraph{Best-of-N Sampling.}

This action addresses the high stochasticity inherent in T2I generation, where multiple plausible images may arise from the same prompt. To reduce variance, the agent produces a set of $N$ candidate images:
\begin{equation}
{ I_{t+1}^{1}, \dots, I_{t+1}^{N} } = g(P_t, I_t),
\end{equation}
and the unified multimodal model assigns each candidate an alignment score:
\begin{equation}
\alpha_t^{k} = \mathrm{Eval}(I_{t+1}^{k}, P_t),
\end{equation}
reflecting semantic consistency and visual quality.
The final selected output is:
\begin{equation}
I_{t+1} = I_{t+1}^{k}, \quad k = \arg\max_k \alpha_t^{k}.
\end{equation}
This sampling strategy stabilizes generation quality and significantly mitigates randomness caused by diffusion or autoregressive sampling.

\section{Detailed Policy Controller Decision}\label{sec:policy_decision}
We present a case study that illustrates the step-by-step decisions made by the policy controller. While Section~\ref{sec:quality} provides a partial walkthrough of how \ours operates on a given example, here we provide the complete decision trajectory, including all intermediate actions, observations, and updates performed by the agent throughout the entire process.

\ours decides to invoke the following actions in order: \textit{Naive Generation}, \textit{Prompt Refinement}, \textit{Image Detail Refinement}, \textit{Prompt Enhancement}, and \textit{Best-of-N Sampling}. Here are the detailed reasons for selecting these actions:

\begin{figure}[h]
    \centering
    \begin{tcolorbox}[
        enhanced,
        colback=green!2!white,
        colframe=green!45!black,
        title=\textbf{Detailed Reasoning for First-step Decision},
        coltitle=white,
        fonttitle=\bfseries\sffamily,
        arc=3mm,
        boxrule=0.8pt,
        left=10pt, right=10pt, top=8pt, bottom=8pt,
        fontupper=\footnotesize\sffamily, 
    ]
    Since no image exists yet and the prompt ``Moldy bread'' is clear and straightforward, it aligns with the first step selection criteria. Using \textbackslash boxed\{naive$\_$generation\} allows for a simple generation at the beginning.
    \end{tcolorbox}
\end{figure}

\begin{figure}[!t]
    \centering
    \begin{tcolorbox}[
        enhanced,
        colback=green!2!white,
        colframe=green!45!black,
        title=\textbf{Detailed Reasoning for Second-step Decision},
        coltitle=white,
        fonttitle=\bfseries\sffamily,
        arc=3mm,
        boxrule=0.8pt,
        left=10pt, right=10pt, top=8pt, bottom=8pt,
        fontupper=\footnotesize\sffamily, 
    ]
    The current image does not match the prompt as it requires a bread with a greenish or blackish mold, but the image provided does not clearly show the mold. The current image may be unsatisfactory or incomplete. Therefore, the best course of action is to use \textbackslash boxed\{prompt$\_$refinement\} to analyze the current image and refine the prompt before regenerating.
    \end{tcolorbox}
\end{figure}

\begin{figure}[!t]
    \centering
    \begin{tcolorbox}[
        enhanced,
        colback=green!2!white,
        colframe=green!45!black,
        title=\textbf{Detailed Reasoning for Third-step Decision},
        coltitle=white,
        fonttitle=\bfseries\sffamily,
        arc=3mm,
        boxrule=0.8pt,
        left=10pt, right=10pt, top=8pt, bottom=8pt,
        fontupper=\footnotesize\sffamily, 
    ]
    The current image does not completely match the prompt as it requires a bread with a greenish or blackish mold, but the image provided just shows a little mold. Therefore, the most appropriate action to use is \textbackslash boxed\{image$\_$detail$\_$refinement\} to generate editing instructions based on the prompt and image, then edit the image accordingly to improve its accuracy and alignment with the prompt.
    \end{tcolorbox}
\end{figure}

\begin{figure}[!t]
    \centering
    \begin{tcolorbox}[
        enhanced,
        colback=green!2!white,
        colframe=green!45!black,
        title=\textbf{Detailed Reasoning for Forth-step Decision},
        coltitle=white,
        fonttitle=\bfseries\sffamily,
        arc=3mm,
        boxrule=0.8pt,
        left=10pt, right=10pt, top=8pt, bottom=8pt,
        fontupper=\footnotesize\sffamily, 
    ]
   The current image is unsatisfactory, and the prompt is not clear and specific. Therefore, the most appropriate action to use is \textbackslash boxed\{prompt$\_$enhancement\} to generate the image based on the detailed prompt provided.
    \end{tcolorbox}
\end{figure}

\begin{figure}[!t]
    \centering
    \begin{tcolorbox}[
        enhanced,
        colback=green!2!white,
        colframe=green!45!black,
        title=\textbf{Detailed Reasoning for Fifth-step Decision},
        coltitle=white,
        fonttitle=\bfseries\sffamily,
        arc=3mm,
        boxrule=0.8pt,
        left=10pt, right=10pt, top=8pt, bottom=8pt,
        fontupper=\footnotesize\sffamily, 
    ]
    The current prompt is already clear, specific, and semantically well-aligned with the intended output. However, text-to-image generation inherently exhibits noticeable randomness, producing visually diverse results even under identical conditions. To mitigate this stochastic variation and ensure a more reliable outcome, the most appropriate action to use is \textbackslash boxed \{best$\_$of$\_$N$\_$sampling\}, which generates multiple candidates from the same prompt and selects the one with the highest visual–semantic alignment.
    \end{tcolorbox}
\end{figure}

All outputs are constrained to be enclosed in \texttt{\textbackslash boxed\{\}}. We then extract the action name using regular expression matching.

\section{Detailed Prompts for Each Action}\label{sec:detailed_prompt_for_each_action}

We present the detailed prompts used for each action in \ours in the following figures. Additionally, the actions \textit{Naive Generation/Editing} and \textit{STOP} directly use the current prompt and therefore do not require additional prompting templates.

\begin{figure}[!h]
    \centering
    \begin{tcolorbox}[
        enhanced,
        colback=green!2!white,
        colframe=green!45!black,
        title=\textbf{Prompt for action \textit{Prompt Enhancement with CoT}},
        coltitle=white,
        fonttitle=\bfseries\sffamily,
        arc=3mm,
        boxrule=0.8pt,
        left=10pt, right=10pt, top=8pt, bottom=8pt,
        fontupper=\footnotesize\sffamily, 
    ]
    \vspace{-2mm}
    You should first think about the planning process in the mind and then generate the image.  The planning process is enclosed within <think> </think> tags, i.e. <think> planning process here </think> image here
    \vspace{-2mm}
    \end{tcolorbox}
\end{figure}

\begin{figure}[!h]
    \centering
    \begin{tcolorbox}[
        enhanced,
        colback=green!2!white,
        colframe=green!45!black,
        title=\textbf{Prompt for action \textit{Prompt Revision Based on the Generated/Edited Image}},
        coltitle=white,
        fonttitle=\bfseries\sffamily,
        arc=3mm,
        boxrule=0.8pt,
        left=10pt, right=10pt, top=8pt, bottom=8pt,
        fontupper=\footnotesize\sffamily, 
    ]
    \vspace{-2mm}
You are an expert in refining and improving image generation prompts while keeping the original intent clear and accurate.\\

**Original Prompt**: "\{prompt\}"\\
**Your Goal**:\\
Review whether the generated image fully captures the meaning, atmosphere, and important visual elements described in the prompt. If not, slightly refine the prompt to make it clearer, more vivid, and easier for an image generation model to understand — while keeping the original intent intact.\\

\#\#\# GUIDELINES\\
1. Keep the **core idea and theme** exactly the same (e.g., Christmas, festival, traditional culture, etc.).\\
2. You may **enhance clarity or atmosphere** by adding descriptive details (e.g., “festive decorations”, “warm lighting”, “crowded street market”).\\
3. Do **not** introduce unrelated content or change the main subject.\\
4. Focus on making the prompt **more visually specific and expressive**.\\
5. If the generated image already matches the prompt well → output the original prompt unchanged.\\

\#\#\# CHECKLIST
- Does the image clearly reflect the main subject or theme?\\
- Is the mood, setting, or cultural context consistent with the prompt?\\
- Could adding small descriptive cues improve visual alignment?\\

**Output Format:**\\
\textbackslash boxed\{[Your final revised prompt here]\}``\\

Analyze the image and output the refined prompt:
    \vspace{-2mm}
    \end{tcolorbox}
\end{figure}

\begin{figure}[!t]
    \centering
    \begin{tcolorbox}[
        enhanced,
        colback=green!2!white,
        colframe=green!45!black,
        title=\textbf{Prompt for action \textit{Image Detail Refinement}},
        coltitle=white,
        fonttitle=\bfseries\sffamily,
        arc=3mm,
        boxrule=0.8pt,
        left=10pt, right=10pt, top=8pt, bottom=8pt,
        fontupper=\footnotesize\sffamily, 
    ]
    \vspace{-2mm}
    You are an expert image editor. Your task is to analyze the current image and provide details editing instructions to make the edited image better match the original prompt.\\

**Original Prompt**: "\{prompt\}"\\

\#\#\# Your Task: \\
1. Compare the current image with the original prompt\\
2. Provide minimal and specific editing instructions for following image editing.\\

\#\#\# Analysis Framework:\\
- **Colors**: Are the colors correct? (e.g., if prompt says "blue apple", is the apple blue?)\\
- **Objects**: Are all mentioned objects present and clearly visible?\\
- **Attributes**: Do sizes, shapes, positions match the description?\\
- **Quality**: Any artifacts, blur, or unclear elements?\\

\#\#\# Assessment Criteria:\\
- Does the image accurately represent the prompt?\\
- Are all key elements present and correctly rendered?\\
- Is the overall quality acceptable?\\
- Would editing significantly improve the result?\\

\#\#\# CRITICAL RULES:\\
- NEVER change the original prompt's colors/objects - only provide editing instructions\\
- Keep unusual colors (brown orange, blue apple, red sheep) - they are intentional\\
- Focus on what needs to be EDITED, not what needs to be generated\\
- Be conservative - only suggest editing if it would genuinely improve the result\\

\#\#\# Output Format:\\
\textbackslash boxed\{[Your editing instructions here]\}\\

Analyze the image and output the editing instructions:\\
    \vspace{-2mm}
    \end{tcolorbox}
\end{figure}

\begin{figure}[!t]
    \centering
    \begin{tcolorbox}[
        enhanced,
        colback=green!2!white,
        colframe=green!45!black,
        title=\textbf{Prompt for action \textit{Best-of-N Sampling}},
        coltitle=white,
        fonttitle=\bfseries\sffamily,
        arc=3mm,
        boxrule=0.8pt,
        left=10pt, right=10pt, top=8pt, bottom=8pt,
        fontupper=\footnotesize\sffamily, 
    ]
    \vspace{-2mm}
You are an expert evaluator of generated images. Your task is to evaluate how well this image matches the given prompt with high precision.\\

**Original Prompt**: "\{prompt\}"\\

\#\#\# Evaluation Framework:\\
- **Attribute Fidelity**: How accurately does the image represent the specific attributes mentioned in the prompt (colors, shapes, textures, objects, etc.)?\\
- **Completeness**: Are all key elements from the prompt present and clearly visible?\\
- **Quality**: Is the image clear, well-composed, and free from obvious artifacts or distortions?\\
- **Consistency**: Do all elements work together coherently and realistically?\\
\\
\#\#\# Instructions:\\
- Examine the image systematically, comparing each aspect mentioned in the prompt\\
- Pay attention to both major elements and subtle details\\
- Consider the overall visual coherence and naturalness\\
- Be critical but fair in your assessment\\

\#\#\# Scoring Scale (0-10):\\
- **10 (Perfect)**: Exceptional match with flawless representation of all prompt elements\\
- **9 (Excellent)**: Nearly perfect with only imperceptible minor deviations\\
- **8 (Very Good)**: Strong match with minor, non-distracting imperfections\\
- **7 (Good)**: Good match with some noticeable but acceptable issues\\
- **6 (Fair)**: Generally matches but has several noticeable problems\\
- **5 (Average)**: Partial match with significant issues affecting key attributes\\
- **4 (Below Average)**: Poor representation with major problems or missing elements\\
- **3 (Poor)**: Serious issues, only vaguely resembles the prompt\\
- **2 (Very Poor)**: Major problems, barely related to the prompt\\
- **1 (Terrible)**: Severely flawed, hardly represents the prompt\\
- **0 (Failure)**: Complete failure or incomprehensible\\

\#\#\# Your Response:\\
Reasoning: [detailed analysis of how well each key element is represented]\\
\textbackslash boxed\{[score 0-10]\}\\
Your Answer:
    \vspace{-2mm}
    \end{tcolorbox}
\end{figure}

\section{Detailed Experimental Results}
We provide the detailed experimental results corresponding to Figure~\ref{fig:wise} in Table~\ref{tab:wise}.

\begin{table*}[t]
  \centering
  \small
    \caption{Experimental results on \textbf{WISE}~\cite{niu2025wise}. Vanilla represents the backbone model, while \ours denotes our agent built upon this model.}

  \setlength{\tabcolsep}{3pt}
\resizebox{\textwidth}{!}{

    \begin{tabular}{lcccccccc}
    \hline
    
    \hline
      Types & Model & \multicolumn{1}{l}{ Cultural} & \multicolumn{1}{l}{Time} & \multicolumn{1}{l}{Space} & \multicolumn{1}{l}{Biology} & \multicolumn{1}{l}{Physics} & \multicolumn{1}{l}{Chemistry} & \multicolumn{1}{l}{Overall}  \bigstrut[t]\\
    \hline
    
   \multirow{6}{*}{Gen} & SDv1.5 & 0.34 & 0.35 & 0.32 & 0.28 & 0.29 & 0.21 & 0.32   \\
    & SDXL & 0.43 & 0.48 & 0.47 & 0.44 & 0.45 & 0.27 & 0.43 \\
    & SD3.5-large & 0.44 & 0.50 & 0.58 & 0.44 & 0.52 & 0.31 & 0.46 \\
    & PixArt-Alpha &  0.45 & 0.50 & 0.48 & 0.49 & 0.56 & 0.34 & 0.47 \\
    & playground-v2.5 & 0.49 & 0.58 & 0.55 & 0.43 & 0.48 & 0.33 & 0.49 \\

     & FLUX.1-dev & 0.48 & 0.58 & 0.62 & 0.42 & 0.51 & 0.35 & 0.50 \\

    \hline

    \multirow{6}{*}{\textit{Unified}}
    & Janus & 0.16 & 0.26 & 0.35 & 0.28 & 0.30 & 0.14 & 0.23 \\
    & VILA-U & 0.26 & 0.33 & 0.37 & 0.35 & 0.39 & 0.23 & 0.31 \\
    & Show-o & 0.28 & 0.40 & 0.48 & 0.30 & 0.46 & 0.30 & 0.35 \\
    & Emu3 & 0.34 & 0.45 & 0.48 & 0.41 & 0.45 & 0.27 & 0.39 \\ 
    & MetaQuery-XL & 0.56 & 0.55 & 0.62 & 0.49 & 0.63 & 0.41 & 0.55 \\
    & GPT-4o & 0.81 & 0.71 & 0.89 & 0.83 & 0.79 & 0.74 & 0.80 \\
    \hline

    \multirow{3}{*}{\textit{Bagel}}
    & Vanilla & 0.44 &
    0.55 & 0.68 & 0.44 & 0.60 & 0.39 & 0.52 \\

    & Self-Rewrite  & \textbf{0.76} & \textbf{0.69} & \textbf{0.75} & \textbf{0.65} & \textbf{0.75}  &\textbf{0.58 }& \textbf{0.70} \\
    
    & \ours & {0.63} & {0.63}  & {0.72} & {0.59} & {0.69} & {0.53} & {0.63}\\

    \hline

    \multirow{3}{*}{\textit{Janus-Pro-7B}}
    & Vanilla & 0.30 & 0.37 & 0.49 & 0.36 & 0.42 & 0.26 & 0.35 \\

    & Self-Rewrite & 0.21 & 0.24 & 0.26 & 0.18 & 0.20 & 0.16 & 0.21  \\
    
    & \ours & \textbf{0.44} & \textbf{0.46}  & \textbf{0.55}  & \textbf{0.45}  & \textbf{0.50} & \textbf{0.27}  & \textbf{0.44}\\
    
    \hline

    \hline
    \end{tabular}}
  \label{tab:wise}

\end{table*}

\section{Benchmarks and Baselines}\label{sec:Detailed_Benchmarks_Baselines}

\paragraph{Benchmarks.}
We evaluate \ours on multiple benchmarks, including 3 benchmarks on image generation and 4 benchmarks for image editing tasks. For image generation, we use R2I-Bench~\cite{chen2025r2i}, which is designed to rigorously assess reasoning-driven T2I generation; WISE~\cite{niu2025wise}, which covers six categories of image generation scenarios; and T2I-ReasonBench~\cite{sun2025t2i}, which assesses the models’ reasoning ability in generative tasks. For image editing, we adopt GEdit-Bench~\cite{liu2504step1x}, which contains both Chinese and English instruction-based editing tasks; RISEBench~\cite{zhao2025envisioning}, which focuses on reasoning-informed visual editing across diverse reasoning types; KRIS-Bench~\cite{wu2025kris}, which evaluates reasoning capabilities over factual, conceptual, and procedural knowledge; and ImgEdit-Bench~\cite{ye2025imgedit}, which is used to evaluate image editing performance in terms of instruction
adherence, editing quality, and detail preservation.

\paragraph{Baselines.} For image generation, we employ SDv1.5~\cite{rombach2022high}, SD3-Medium~\cite{rombach2022high}, SDXL~\cite{podell2023sdxl}, SD3.5-Medium~\cite{esser2024scaling}, SD3.5-Large~\cite{esser2024scaling}, PixArt-Alpha~\cite{chen2024pixart}, FLUX.1-Dev~\cite{flux2024}, Sana-1.5~\cite{xie2025sana}, Lumina-T2I~\cite{qin2025lumina}, LLM4GEN$_{SD1.5}$~\cite{liu2025llm4gen}, ELLA$_{SD1.5}$~\cite{hu2024ella}, LlamaGen~\cite{sun2024autoregressive}, DALL-E-3~\cite{ma2025learning}, gpt-image-1~\cite{hurst2024gpt}, Omnigen~\cite{xiao2025omnigen}, FLUX.1-schnell~\cite{flux2024}, and Playground-v2.5~\cite{li2024playground},  as generation-based models. We also include Janus~\cite{wu2025janus}, VILA-U~\cite{wu2024vila}, Show-o~\cite{xie2024show}, Janus-Pro-7B~\cite{chen2025janus}, Emu3~\cite{wang2024emu3}, Lumina-Image 2.0~\cite{qin2025lumina}, show-o~\cite{xie2024show}, GoT~\cite{fang2025got} and MetaQuery-XL~\cite{pan2025transfer} as unified multimodal models.

For image editing, we evaluate both private and open baselines. The private baselines include Gemini-2.0~\cite{google2025_gemini2_image}, Doubao~\cite{shi2411seededit}, GPT-4o~\cite{openai2025gpt4o_image}, and Gemini-2.0-Flash-pre~\cite{team2023gemini}. The open baselines include Step1X-Edit~\cite{liu2025step1x}, Instruct-Pix2Pix~\cite{brooks2023instructpix2pix}, MagicBrush~\cite{zhang2023magicbrush}, AnyEdit~\cite{yu2025anyedit}, OmniGen~\cite{xiao2025omnigen}, EMU2~\cite{sun2024generative}, Qwen-Image-Edit~\cite{wu2025qwenimagetechnicalreport}, FLUX.1-Kontext-Dev~\cite{labs2025flux1kontextflowmatching}, Ovis-U1~\cite{wang2025ovisu1}, and Seedream-4.0~\cite{seedream2509seedream}. 

Meanwhile, we also compare \ours with the backbone unified model and a prompt \textit{self-rewrite} baseline implemented using the same backbone model. The prompt for the self-rewrite baseline is identical to the official prompt used in Bagel, which we present in Figure~\ref{fig:self-rewrite-prompt}.

\begin{figure}[t]
    \centering
    \begin{tcolorbox}[
        enhanced,
        colback=green!2!white,
        colframe=green!45!black,
        title=\textbf{Prompt for Self-rewrite Baseline},
        coltitle=white,
        fonttitle=\bfseries\sffamily,
        arc=3mm,
        boxrule=0.8pt,
        left=10pt, right=10pt, top=8pt, bottom=8pt,
        fontupper=\footnotesize\sffamily, 
    ]
    You should first think about the planning process in the mind and then generate the image. The planning process is enclosed within <think> </think> tags, i.e. <think> planning process here </think> image here.
    \vspace{1mm}

    \end{tcolorbox}
    \vspace{-2mm}
    \caption{Detailed prompt for self-rewrite baseline.}
    \label{fig:self-rewrite-prompt}
    \vspace{-1mm}
\end{figure}

\section{Additional Experiments on KRISBench}

Due to space limitations in the main paper and the lack of suitable baselines, we place the detailed results on KRISBench in this section. Specifically, KRISBench contains several subcategories involving image editing tasks that require transforming multiple images into a single output image. However, Bagel does not support such operations. Therefore, we remove the corresponding subcategories and only compare our method with the Vanilla baseline under the remaining settings.

The detailed results are presented in Table~\ref{tab:krise_edit}. The results consistently demonstrate that \ours outperforms the Vanilla baseline across the evaluated categories.

\begin{table*}
  \centering
    \caption{
    Experimental results on \textbf{KRISBench}~\cite{wu2025kris}.}
    \begin{tabular}{cccccccc}
    \hline
    
    \hline

     \multirow{2}{*}{Method} & \multicolumn{2}{c}{Factual} & \multicolumn{2}{c}{Conceptual} & \multicolumn{2}{c}{Procedural} & \multirow{2}{*}{Overall} \\
     \cline{2-7}
        
     & \multicolumn{1}{c}{AP} & \multicolumn{1}{c}{SP} & \multicolumn{1}{c}{SS} & \multicolumn{1}{c}{NS} & \multicolumn{1}{c}{LP} & \multicolumn{1}{c}{ID}  \bigstrut[t] \\
    \hline

     Vanilla & 73.27 & 67.17 & 61.10 & 62.36 & 54.38 & 64.77 & 63.16 \\
     ~\ours & \textbf{74.39} & \textbf{71.15} & \textbf{69.80} & \textbf{66.03} &\textbf{54.83} & \textbf{68.78} & \textbf{67.13} \\
     
    \hline

    \hline

    \end{tabular}

  \label{tab:krise_edit}
  
\end{table*}

\end{document}